\pdfoutput=1
\documentclass[11pt]{article}

\usepackage{EMNLP2023}

\usepackage{times}

\usepackage{latexsym}
\usepackage{booktabs}
\usepackage{multirow}
\usepackage{amsmath}
\usepackage{url}
\usepackage{graphicx}
\usepackage{ulem}
\usepackage{algorithm}
\usepackage{algorithmic}
\usepackage{enumitem}
\usepackage{amsmath,lipsum}
\usepackage{mathrsfs}
\usepackage{amssymb}

\usepackage[T1]{fontenc}
\usepackage[utf8]{inputenc}

\usepackage{microtype}

\usepackage{inconsolata}

\title{Mitigating Large Language Model Hallucination with Faithful Finetuning}

\author{
  Minda Hu$^{1}$\hspace{0.3cm}
  Bowei He$^2$\hspace{0.3cm}
  Yufei Wang$^3$\hspace{0.3cm}
  Liangyou Li$^3$\hspace{0.3cm}
  Chen Ma$^2$\hspace{0.3cm}
  Irwin King$^1$\\
  $^1$Dept. of Computer Science \& Engineering, The Chinese University of Hong Kong\\
  $^2$Department of Computer Science, City University of Hong Kong\\
  $^3$Huawei Noah’s Ark Lab\\
  \texttt{\{mindahu21, king\}@cse.cuhk.edu.hk}
  }

\begin{document}
\maketitle
\begin{abstract}
Large language models (LLMs) have demonstrated remarkable performance on various natural language processing tasks. However, they are prone to generating fluent yet untruthful responses, known as "hallucinations". Hallucinations can lead to the spread of misinformation and cause harm in critical applications. Mitigating hallucinations is challenging as they arise from multiple complex factors like overconfidence and lacking knowledge, especially in Question Answering (QA) tasks. %such as noisy data, model overconfidence, lack of knowledge, and the generation process itself. 
Recent efforts have attempted to address this issue through representation editing and decoding algorithms without major structural changes or retraining. However, these approaches either implicitly edit LLMs' behavior in latent space or suppress the tendency to output unfaithful results during decoding instead of explicitly modeling on hallucination. In this work, we introduce \textbf{F}aithful \textbf{F}inetuning (\textbf{F2}), a novel method that explicitly models the process of faithful question answering through carefully designed loss functions during fine-tuning. We conduct extensive experiments on popular datasets and demonstrate that F2 achieves significant improvements over vanilla models and baselines. %Furthermore, we show that our method is empirically orthogonal to current state-of-the-art manipulation methods, further enhancing their strong performance on TruthfulQA. Our findings emphasize the effectiveness of explicit loss design in improving LLM truthfulness and pave the way for more reliable and trustworthy language models in real-world applications.
\end{abstract}

\section{Introduction}

Large language models (LLMs) have emerged as transformative tools in the field of natural language processing (NLP), showcasing unparalleled proficiency across a diverse array of tasks~\citep{achiam2023gpt, touvron2023llama, zhang2023bayling}. Despite their impressive capabilities, LLMs are occasionally prone to generating responses that, while coherent and seemingly in line with given instructions, diverge from the truth - a phenomenon aptly termed as "hallucinations"~\citep{JiLFYSXIBMF23}. Such inaccuracies not only compromise the reliability of LLMs but also pose significant challenges for their application in critical domains.

The root causes of these hallucinations remain a subject of ongoing investigation. Factors contributing to this issue may include the models' overreliance on their own outputs, the preference for generating fluent text at the expense of accuracy, and the intrinsic uncertainties associated with the knowledge amassed during their training phase. The implications of hallucinations are profound, potentially facilitating the dissemination of misinformation, diminishing trust in AI technologies, and causing detrimental effects when employed in sensitive decision-making contexts.

In addressing this challenge, specially within the context of Question Answering (QA), the academic community has dedicated considerable effort towards devising strategies to curtail the occurrence of hallucinations. Prior studies~\citep{QiuZKPC23, chuang23, kai2024sh2, li2024inference} have illustrated that it is feasible to significantly mitigate hallucinations without necessitating extensive structural modifications or comprehensive retraining of the models. Nevertheless, many of these approaches either implicitly adjust the behavior of LLMs within the latent space or aim to suppress the tendency of outputting unfaithful results during the decoding phase, essentially treating LLMs as opaque entities.

Contrary to these approaches, our research introduces a novel approach that emphasizes the fidelity of responses through the meticulous design of explicit loss functions during the fine-tuning process, named as \textbf{F}aithful \textbf{F}inetuning (\textbf{F2}). We first decompose the conventional QA objective into two explicit sub-objectives: internal fact retrieval and fact-grounded QA, thus informing the LLM to effectively leverage their internal knowledge for faithful answers. Then, we design a targeted fine-tuning approach to guide the model focusing on the hotspots identified by the entity-based and attention-based heuristics within the retrieved fact spans. Additionally, to minimize the occurrence of model hallucinations, we select the hallucination-prone layers in the LLM structure and specifically conduct the fine-tuning on them. Empirical evaluations conducted on the TruthfulQA~\citep{LinHE22} and FACTOR\citep{MuhlgayRMLRBALSS24} datasets, which are widely employed in the study of LLM hallucinations~\citep{zhang2024truthx, li2024inference}, reveal that our method achieves noticeable improvement compared to the vanilla models. In addition, our method is empirically proven to be orthogonal to the current state-of-the-art manipulation methods, further boosting their already strong performance on such benchmarks. Our findings emphasize the effectiveness of explicit loss design in improving LLM truthfulness and pave the way for more reliable and trustworthy language models in real-world applications.

\begin{figure*}[t!]
    \centering
    \includegraphics[width=0.9\textwidth]{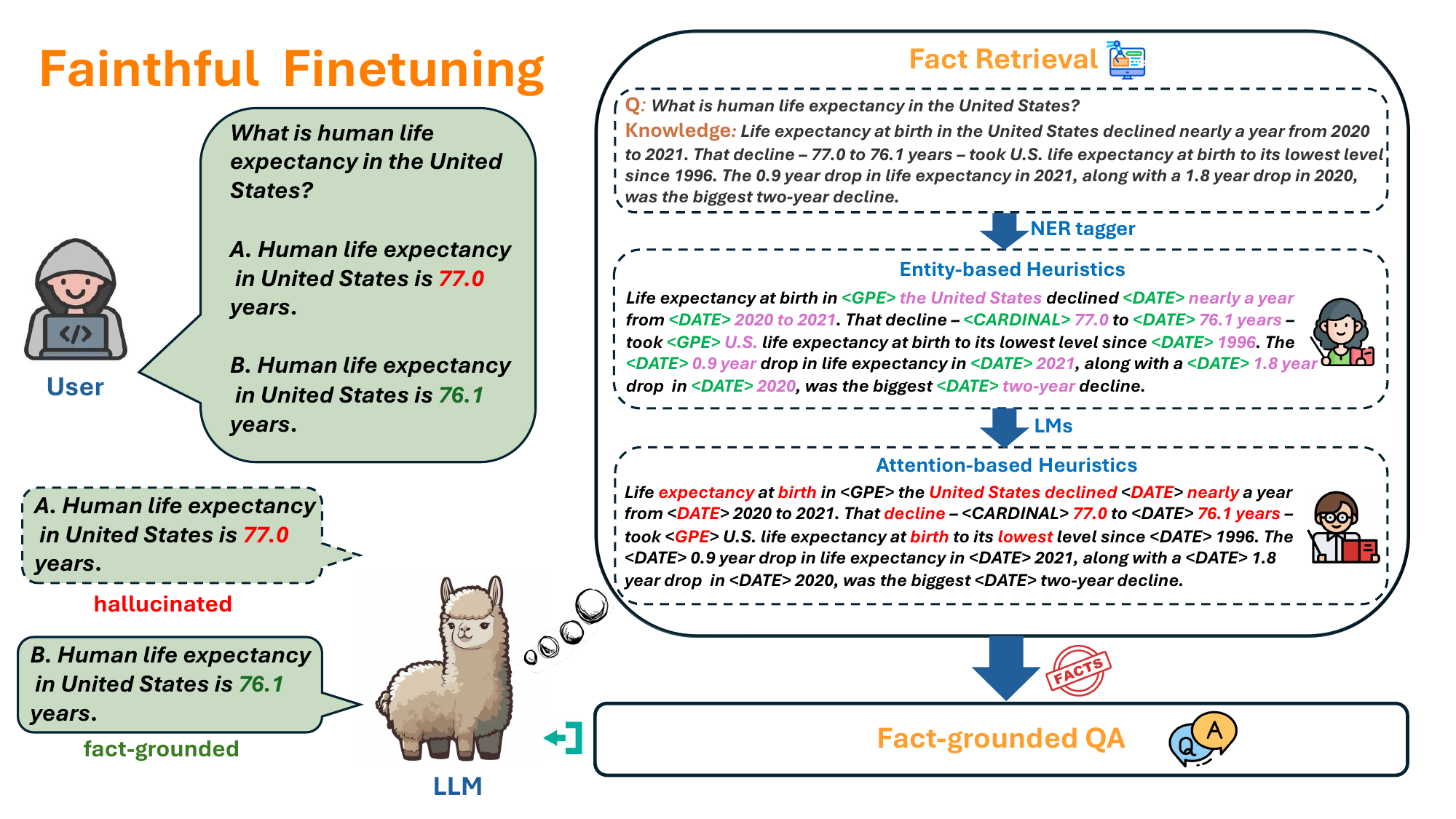}
    \caption{Overview of the Faithful Finetuning method.}
    \label{fig:f2-overview}
\end{figure*}

\section{Related Works}
\paragraph{Hallucination in Generative Models}
As the adoption of text generative models in various applications ranging from automated summarization~\citep{summa2021}, machine translation~\citep{DaleVBC23}, question answering~\citep{SadatZLAGWMP023}, to dialogue systems~\citep{DziriMYZR22} has surged, the hallucination phenomenon has drawn considerable attention within the NLP community. Hallucinations refers to the generation of content that is either unfaithful to the input data or devoid of factual grounding~\citep{MaynezNBM20, leiyu24}. This phenomenon poses challenges to the reliability and applicability of such models, especially in the critical domains requiring high levels of accuracy and trustworthiness. Generally, the hallucination in text generation can be categorized into two types: \textit{extrinsic hallcination} and \textit{intrinsic hallucination}~\citep{JiLFYSXIBMF23}. The former one occurs when the generated text includes factual inaccuracies or entities not present in the input, while the second one is irrelevant and does not address the prompt or the task at hand though factually correct.

\paragraph{Mitigating LLM Hallucination}
Though recent works have recognized the severity of LLM hallucination problem. many previous works still only focus on detecting such hallunications~\citep{WangYHZH23, xiangchen24}, while neglecting the importance of mitigating them during the text generation process. Existing mitigation approaches~\citep{mitigate2024} mainly consist of two categories: \textit{prompt engineering} and \textit{developing models}. The \textit{prompt engineering} here refers to designing the input instructions to get the best output possible while not touching the model architectures or parameters, exhibiting the advantages of being lightweight and fast. The common apporachs in this category include retrieval augmented generation~\citep{varshney23, peng23}, self refinement through feedback and reasoning~\citep{Dhuliawala23, ziweiji23}, and prompt tuning~\citep{ChengHBZLW0WDZ23, jones23}. Instead of editing the LLM as a black box, the \textit{developing models} methods focus on developing or modifying model architecture or parameters to pertinently address the hallucination challenge. Introducing new decoding strategy~\citep{weijia23, chuang23}, utilization of knowledge graph~\citep{zihan23, BayatQHSBKWI023}, introducing faithfullness-based loss function~\citep{YoonYYKY22, QiuZKPC23}, and supervised finetuning~\citep{Katherine23, Evgeniia23} are four main schemes in this category. In this paper, we also follow this type of approach and explore how to mitigate the hallucination by eliminating some deficiencies and flaws in the LLM itself. So far as we know, we are the first to introduce the idea of a heuristic-based weighting strategy and task decomposition into the field of hallucination mitigation.
\section{Method}
To improve the faithfulness of Question Answering (QA) models, we propose a multi-task training approach, \textsc{Faithful Finetuning (F2)}. This approach decomposes the QA objective into two explicit sub-objectives:

\begin{enumerate}
    \item \textbf{Internal Fact Retrieval}: This objective trains the model to effectively retrieve and leverage its internal knowledge to produce faithful answers.
    \item \textbf{Fact-grounded QA}: This objective trains the model to provide answers that are grounded in factual information. 
\end{enumerate}

The key insight behind F2 is that by explicitly training the model on these sub-objectives, we can enhance its ability to access and utilize its internal knowledge, rather than relying on potentially unreliable hallucations.

Furthermore, based on observations about LLMs' behavior to hallucinate from the perspective of output probability and network structure, F2 leverages weighted objectives and targeted fine-tuning on hotspots including hallucination-prone spans and layers. These weighted objectives underline the spans that LLMs are prone to hallucinate and strengthen the LLMs' capability to retrieve reliable and crucial information needed for accurate QA tasks.

By decomposing the QA task and incorporating targeted training on hallucination-prone areas during fact retrieval, F2 aims to produce QA models that are more faithful and grounded in factual information.
\subsection{Multi-objective Decomposition for Faithful QA}
Conventionally speaking, the objective of the QA task is to directly predict answer sequence $a$ based on question $q$. To strictly regularize LMs to answer faithfully, we propose decomposing the QA task by adding two explicit sub-objectives: Internal Fact Retrieval and Fact-grounded QA.
\subsubsection{Vanilla QA Objective}
In the conventional QA Objective, the conditional probability of $a$ given $q$ is increased by optimizing the cross entropy loss $\mathcal{L}_{QA}(\phi)$:
    \begin{equation}
        \begin{aligned}
            & \mathcal{L}_{QA}(\phi)= \\
            & -\mathbb{E}_{(k, q, a) \sim D_{FQA}}\left[\log \tau_\phi(a \mid q)\right]
        \end{aligned}
    \end{equation},
where the function $\tau_\phi(a \mid q)$ represents the probability distribution generated by the auto-regressive LMs with parameters $\phi$, predicting the answer $a$  based on the question prompt $q$. Following the implementation of TruthfulQA, the prompt template for vanilla QA is formatted below. The complete question prompt is illustrated in Table~\ref{tab:complete_question_prompt}.

\begin{table}[!ht]
%\small
    \centering
    % \colorbox[HTML]{edf2fb}{
    \colorbox{orange!8}{
    \begin{tabular}{@{}p{7.2cm}}
    Q: What is human life expectancy in the United States?\\
    A: Human life expectancy in the United States is 78 years.\\
    ......\\
    Q: \textbf{\{question\}}\\
    A: \textbf{\{answer\}}
    \end{tabular}}
    %\vspace{0.2cm}
    %\caption{Question Prompt}
    \label{tab:question_prompt}
\end{table}
%\vspace{-0.5cm}
\vspace{-3mm}
%\begin{enumerate}
\subsubsection{Fact Retrieval Objective}
This objective is designed to split the loss function to improve internal knowledge representation, enhancing LMs' capability to access their internal memory and retrieve relevant and factual knowledge  $k$ given question $q$ in a self-contained manner, to enable LMs to provide more accurate and informative responses without relying on external resources:
    \begin{equation}
        \begin{aligned}
            & \mathcal{L}_{R}(\phi)= -\mathbb{E}_{(k, q, a) \sim D_{FQA}}\left[\log \tau_\phi(k \mid q)\right]
        \end{aligned}
    \end{equation}
Given a QA training dataset $(k, q, a)$ with grounding fact $k$, LMs are required to produce $k$ given only $q$. The prompt for fact retrieval task is simply constructed as \textit{Q: \textbf{\{question\}} Knowledge:\textbf{\{fact\}}}.
\subsubsection{Fact-grounded QA Objective}
% The knowledge-guided QA objective is devised to encourage models to produce: responses $a$ grounded in the retrieved knowledge $k$:
The Fact-grounded Question-Answering (FQA) objective is specifically designed to encourage language models to generate responses $a$ that are firmly grounded in the fact $k$ retrieved from their internal memory. By incorporating this objective, the model is incentivized to carefully consider and utilize the most relevant and factual information available within its internal memory when formulating answers. This approach aims to ensure that the model's responses are not only coherent but supported by the facts. By promoting the integration of retrieved knowledge into the answer-generation process, the FQA objective seeks to enhance the accuracy and reliability of the model's responses to given questions.
    \begin{equation}
        \begin{aligned}
            & \mathcal{L}_{FQA}(\phi)= \\
            & -\mathbb{E}_{(k, q, a) \sim D_{FQA}}\left[\log \tau_\phi(a \mid q, k)\right]
        \end{aligned}
    \end{equation}
%\end{enumerate}
Combined with $\mathcal{L}_{R}(\phi)$, $\mathcal{L}_{R}(\phi) + \mathcal{L}_{FQA}(\phi)$ can effectively optimize the joint probability of generating response $a$ and retrieving knowledge $k$ given question $q$. By simultaneously optimizing these two objectives, the model learns to strengthen the correlation between the retrieved knowledge and the generated answer. The mathematical justification for this approach is provided in Equation \ref{eq:joint_prob}, which demonstrates how the joint optimization of the retrieval and knowledge-guided QA objectives leads to an increased alignment between the retrieved knowledge and the generated responses, ultimately improving the model's ability to provide accurate and well-supported answers to given questions. The prompt for fact retrieval task is simply constructed as \textit{Q: \textbf{\{question\}} Knowledge:\textbf{\{fact\}} A:\textbf{\{answer\}}}.
%can optimize the joint probability of $a$ and $k$ given $q$, increasing the correlation between $a$ and $k$ and further restricting LMs from answering with hallucinated groundings. The mathematical proof is in Eq. \ref{eq:joint_prob}.

%\small
    \begin{equation}
        \label{eq:joint_prob}
        \begin{aligned}
            &\mathcal{L}_{R}(\phi) + \mathcal{L}_{FQA}(\phi) = \\
            -&\mathbb{E}_{(k, q,a) \sim D_{FQA}}\left\{\log \left[\tau_\phi(a \mid q, k) \times \tau_\phi(q \mid k)\right]\right\} \\
            & \propto -\mathbb{E}_{(k, q, a) \sim D_{FQA}}\left[\log \tau_\phi(a, q\mid k)\right] \\
        \end{aligned}
    \end{equation}
\normalsize
%\begin{equation}
%    \begin{aligned}
%        \mathcal{L}_{D}(\phi)= &\mathcal{L}_{retrieval}(\phi) + \mathcal{L}_{FQA}(\phi) +\\
%        & \mathcal{L}_{QA}(\phi)
%    \end{aligned}
%\end{equation}

\subsection{Targeted Training on Hallucination Hotspots}
%To further improve faithfulness, we conduct targeted fine-tuning on hotspots in retrieved fact spans and transformer layers. We identify spans in the training data where the model is prone to hallucinating and up-weight the loss on those spans during training. These spans are located according to entity-based and attention-based heuristics inspired by previous work on hallucination explanation and detection \md{[cite here]} In addition, we select and finetune the top 10 modules most related to hallucination, as measured by probing \md{[cite here]}.
To further enhance the faithfulness of the generated responses, we employ a targeted fine-tuning approach that focuses on hotspots identified within the retrieved fact spans in $\mathcal{L}_{R}(\phi)$, as well as layers in the LLMs. According to previous works, we recognize that certain spans in the training data are more susceptible to model hallucination. To tackle this issue, we assign higher weights to the loss function on these specific spans during the training process. The identification of these critical spans is guided by both entity-based and attention-based heuristics, drawing inspiration from previous research on hallucination explanation and detection~\cite{zhang2023enhancing, WangYHZH23, xiangchen24}. Moreover, we select and fine-tune the top 10 modules that exhibit the strongest association with hallucination, as determined through probing techniques. 
%By concentrating on these key components of the model architecture, we aim to mitigate the generation of unfaithful or inconsistent responses and improve the overall reliability of the model's outputs. 
%This targeted fine-tuning approach enables the model to better align its generated responses with the factual information present in the retrieved knowledge, ultimately enhancing the trustworthiness of the question-answering system.

%\subsubsection{Adding NER Tag \md{change the subsection}}
%\md{[NER Tag Intro Here]}

\subsubsection{Entity-based Heuristics}
%\md{[change this]}Previous work has highlighted the stark disparities in how models and humans assess information: when evaluating the preceding words, the model focuses meticulously on all the feasible choices with different entity types, while humans tend to intuitively include it within a tailored set of candidate words that predominantly consists of terms related to just a few types. This can lead to underconfidence in the model's predictions.

%Prior works (Pagnoni et al., 2021; Kry ́ sci  ́ nski et al., 2020) suggest that entities are the most frequently hallucinated words in text generation. This aligns with the intuition that, when evaluating the veracity of generated results, our primary focus lies on keywords. Weighted Cross Entropy~(WCE) loss is effective in biasing the LM towards generating constraint terms. \md{[cite this]} 

Studies by~\citeauthor{PagnoniBT21} and ~\citeauthor{KryscinskiMXS20} indicate that entities are the most common type of words that are hallucinated or fabricated in text generation tasks. This finding aligns with the intuition that when assessing the truthfulness of generated text, people tend to focus mainly on keywords. To address this issue, we propose to leverage the Weighted Cross Entropy (WCE) loss. WCE has proven effective in encouraging language models to generate words that adhere to specific constraints~\cite{ailem2021encouraging}.

%To address this, we introduce two types of focal heuristics to identify hallucination hotspots: entity-based and attention-based. The entity-based focal heuristic assigns higher weights to entity spans:

To guide models to include these important entities in retrieved internal facts, we modify the aforementioned $\mathcal{L}_{retrieval}(\phi)$ by applying WCE loss during fine-tuning: 

%\small
\begin{equation}
    \label{eq:wce_1}
    \begin{aligned}
         \mathcal{L}^{WCE}(\phi, w, D)& = -\mathbb{E}_{(k, q, a) \sim D}\left[\tau^{WCE}(k \mid q, w)\right],\\
         \mathrm{where}\ \tau^{WCE}&(k \mid q, w)\\
        = \sum^{|k|}_{i=1}&w_i\log p(k_i \mid q, k_1, ..., k_{i-1})
    \end{aligned}
\end{equation}
%\normalsize
, where $w$ is the weight list and $D$ is training set with groundings.

We identify these named entity spans using Spacy\footnote{\url{https://spacy.io/api/entityrecognizer}} and put more weight on the spans of entities:
\begin{equation}
    \label{eq:wce_2}
    \begin{aligned}
         &\mathcal{L}_{E}(\phi) = \mathcal{L}^{WCE}(\phi, w^{ent}, D_{FQA}), \\
         &\mathrm{where}\ w^{ent}_i = \left\{\begin{array}{ll}
         \alpha \mbox{, if } i \in span_{ent}\\
         1 \mbox{, otherwise}
         \end{array}\right.
    \end{aligned}
\end{equation}

For entity spans $span_{ent}$, we only include those in $k$ instead of in The FQA prompt, the details of extracting entity spans are shown in Algorithm~\ref{alg:ent_spans}.
%\textbf{Entity-based Heuristics:}
%\md{[NER Tag Intro Here, talk about underconfidence problem]}

\begin{algorithm}
   \caption{Extracting Entity Spans}
   \label{alg:ent_spans}
\begin{algorithmic}
   \STATE {\bfseries Input:} Question $q$, Sample Fact $k$, Prompt Template for Fact Retrieval $P_{FR}$, NER Tagger $f_{ner}$.
   %\REPEAT
   \STATE $span_{ent}$ = $\emptyset$
   \STATE input = $P_F$($k$, $q$)
   \STATE prompt\_input = $P_{FR}$($q$)
   \STATE ent\_span = $f_{ner}$(input)
   \FOR{$i=0$ {\bfseries to} $\mathrm{|ent\_spans|}$}
   %\REPEAT
   \STATE $\mathrm{tok_{start}}$, $\mathrm{tok_{end}}$ = $\mathrm{ent\_span_{i}}$
   \FOR{$j=\mathrm{tok_{start}}$ {\bfseries to} $\mathrm{tok_{end}}$}
        \IF{$j$ < $\mathrm{|prompt\_input|}$}
            \STATE Add $j$ to Set $span_{ent}$
        \ENDIF
   \ENDFOR
   %\UNTIL{no bug}
   \ENDFOR
   \RETURN $span_{ent}$
   %\UNTIL{$T$ is complete}
\end{algorithmic}
\end{algorithm}

In addition to focusing on entity spans, previous research has revealed significant differences in how language models and humans evaluate information. When assessing the words that come before, models tend to consider various options with diverse entity types. In contrast, humans intuitively narrow down the candidate words to a specific set, primarily consisting of terms related to a limited number of types. This discrepancy can lead to the model's predictions appearing less confident. However, narrowing down a candidate set poses a challenge during the model generation. Inspired from \citeauthor{zhang2023enhancing}, we leverage the in-context learning capability of the LLMs by inserting the named entity type preceding every named entity identified by Spacy. The entity type serves as a generation constraint, enabling us to approximate the ideal candidate set.

At last, we get the final loss design with entity-based heuristics:
\begin{equation}
    \begin{aligned}
         &\mathcal{L}^{tag}_{E}(\phi) = \mathcal{L}^{WCE}(\phi, w^{ent}, D_{FQA}^{tag}), \\
    \end{aligned}
\end{equation}
where $k^{tag}$ in $(k^{tag}, q, a) \in D_{FQA}^{tag}$ are grounding facts with the named entity types inserted, as shown in Figure~\ref{fig:f2-overview}.

%To address this, we introduce two types of focal heuristics to identify hallucination hotspots: entity-based and attention-based. The entity-based focal heuristic assigns higher weights to entity spans:

\subsubsection{Attention-based Heuristics} 

In addition to entity-based heuristics, we construct a weighted graph from the max-pooled attention matrix to include more crucial tokens to LMs and preserve more underlying information contained in the related sample facts, such as the reasoning between entities. The attention-based focal heuristic assigns higher weights to spans with high attention scores:
\begin{equation}
    \begin{aligned}
        & \mathcal{L}_{A}(\phi) = \mathcal{L}^{WCE}(\phi, w^{ent}, D_{FQA}), \\
        & \mathrm{where}\ w^{attn}_i = \left\{\begin{array}{ll}
         \alpha & \mbox{, if } i \in span_{attn}\\
         1 & \mbox{, otherwise}
         \end{array}\right.
    \end{aligned}
\end{equation}
Then we measure the saliency of the related facts by the PageRank algorithm~\cite{rogers2002google}, which is a popular algorithm for ranking the importance of nodes in a graph based on the structure of incoming links. PageRank operates on the premise that the importance of a node is determined not only by the number of links it receives but also by the importance of those linking nodes. Essentially, it assigns a numerical importance to each node within the graph, with higher scores indicating higher importance. This iterative algorithm uses the link structure of the graph to distribute ranking power through the network, allowing us to identify the most salient facts within our attention-based graph. The top-K tokens ranked by PageRank scores are also weighted in the fine-tuning loss.
\begin{algorithm}
   \caption{Extracting Attention Spans}
   \label{alg:attn_spans}
\begin{algorithmic}
   \STATE {\bfseries Input:} Question $q$, Fact $k$, Prompt Template for Fact Retrieval $P_{FR}$, LMs $\phi$, Graph Constructor $G$, Top-K $K$.
   %\REPEAT
   \STATE $span_{attn}$ = $\emptyset$
   \STATE input = $P_{FR}$($k$, $q$)
   \STATE prompt\_input = $P_{FR}$($q$)
   \STATE $A$ = $\phi$(input)
   \STATE $A_{pooled}$ = max\_pooling(attn\_matrix)
   \STATE $g$ = $G$($A_{pooled}$)
   \STATE top\_k\_tokens = page\_rank($g$, $K$)
   \FOR{$i=0$ {\bfseries to} $\mathrm{|top\_k\_tokens|}$}
   %\REPEAT
   \STATE $tok$ = top\_k\_tokens[i]
    \IF{$tok$ < $\mathrm{|prompt\_input|}$}
        \STATE Add $tok$ to set $span_{attn}$
    \ENDIF
   %\UNTIL{no bug}
   \ENDFOR
   \RETURN $span_{attn}$
   %\UNTIL{$T$ is complete}
\end{algorithmic}
\end{algorithm}

In Faithful Finetuning, we merge weights $w^{attn}$ and $w^{ent}$ into $w^{attn \cup ent}$ and jointly regularize the model to retrieve important factual information $k$ related to $q$.

%\small
\begin{equation}
    \begin{aligned}
     & \mathcal{L}_{A \cup E}(\phi) = \mathcal{L}^{WCE}(\phi, w^{attn \cup ent}, D_{FQA}), \\
     & \mathrm{where}\ w^{ent \cup attn}_i = \left\{\begin{array}{ll}
     \alpha & \mbox{, if } i \in span_{attn \cup ent}\\
     1 & \mbox{, otherwise}
     \end{array}\right.
    \end{aligned}
\end{equation}
\normalsize

To validate our proposed entity and attention-based heuristics, we conduct an observation experiment. The results, detailed in Appendix~\ref{sec:observ_detail}, show that our weighting strategies exhibit a stronger correlation with hallucination behavior, as measured by the Spearman coefficient, compared to the average strategy used in $\mathcal{L}_{R}(\phi)$.

Combining all the heuristics mentioned above, we get the final form of fine-tuning objective in Faithful Finetuning:
\begin{equation}
    \begin{aligned}
    \mathcal{L}^{tag}_{F2}(\phi) & = \mathcal{L}^{tag}_{QA}(\phi) + \mathcal{L}^{tag}_{FQA}(\phi) \\
    & + \mathcal{L}^{tag}_{A \cup E}(\phi)
    \end{aligned}
\end{equation}
, where $\mathcal{L}^{tag}_{QA}$ and $\mathcal{L}^{tag}_{FQA}$ are aforementioned loss $\mathcal{L}_{QA}$ and $\mathcal{L}_{FQA}$ using tagged training set $D_{FQA}^{tag}$.

%\subsubsection{Observation}

\subsubsection{Finetuning Hallucination-Prone Layers} 
%To optimize the model for faithful question answering,, we follow the approach proposed in TruthX. We fine-tune only the top 10 modules most related to hallucination, as measured by probing accuracy on the validation set. For instance, for a 32-layer language model, TruthX selects the top 10 modules with the highest probing accuracy out of the total 64 modules (32 attention modules + 32 FFN modules) to edit the model. The structure selection experiment result is presented in Table~\ref{tab:layer_select}, and the implementation details are in the Appendix.
To optimize the model for faithful question answering, we adopt the approach proposed in TruthX~\cite{zhang2024truthx}. This method involves fine-tuning only the top 10 modules most strongly associated with hallucination, as determined by probing accuracy on the validation set. For example, in a 32-layer language model, TruthX selects the top 10 modules with the highest probing accuracy from a total of 64 modules (32 attention modules and 32 FFN modules) for model editing. The structure selection experiment results are presented in Table~\ref{tab:layer_select}.

\begin{table}[!ht]
    \centering
    \small
    %\makebox[\linewidth][c]{
    %resizebox{\textwidth}{16mm}{
    %\setlength{\tabcolsep}{3.3mm}{
    \begin{tabular}{lccc}
    \toprule
         \multirow{2}{*}{Selection Method}  % 92
         & \multicolumn{3}{c}{TruthfulQA~($MC_{max}$)} % 108
        \\\cmidrule(lr){2-4}
         ~%& Expert 
         & MC1 & MC2 & MC3 \\
         \midrule
         %ETHICS &  && 59.9 / 38.2 & && 64.1 / 37.2  & && 81.9 / 67.4\\
         \textsc{All Layers} & 50.31 & 71.85 & 45.30\\
         \textsc{Selected Top-10} &  \textbf{51.29} & \textbf{72.97} & \textbf{45.79}\\
         %& - 
        \bottomrule
        %   GPT4-Justice-v1-yes/no & 0.8621  &  0.5495 &   0.6711 & 0.755 \\
    \end{tabular}
    %}
    \caption{Results of different layer selection strategies. TruthfulQA~($MC_{max}$) reports the highest metrics achieved during the whole fine-tuning process.}
    \vspace{-0.2cm}
    \label{tab:layer_select}
\end{table}
By tackling the issue of hallucination from multiple angles of loss design and fine-tuning, combining this strategy complements our F2 methods, and it can further boost the performance of Faithful Finetuning.

\section{Experiments}
\subsection{Datasets}
We utilize three datasets in our experiments:
\begin{itemize}[leftmargin=*]
    \item \textbf{HaluEval}: This benchmark, known as the Hallucination Evaluation for Large Language Models (HaluEval) \citep{LiCZNW23}, comprises a comprehensive set of 35,000 samples, both hallucinated and normal, designed for the analysis and evaluation of LLMs. It encompasses 5,000 general user queries answered by ChatGPT and an additional 30,000 task-specific instances drawn from areas such as question-answering, knowledge-grounded dialogue, and text summarization. For our F2 approach, we specifically utilize its question-answering subset, which includes 10,000 hallucinated QA samples derived from HotpotQA \citep{Yang0ZBCSM18}, featuring knowledge from Wikipedia, a question, a verified answer from HotpotQA, and a corresponding hallucinated response.
    \item \textbf{TruthfulQA}:  This benchmark \citep{LinHE22}, recognized for measuring the truthfulness of LLMs, contains 817 questions across 38 distinct categories. Our experiments leverage its multiple-choice discrimination tasks as a test set, wherein the LLM is tasked with selecting the correct answer from a set of both accurate and inaccurate options, assessed via multiple-choice accuracy metrics (MC1, MC2, and MC3).
    \item \textbf{FACTOR}: The FACTOR benchmark \citep{MuhlgayRMLRBALSS24}, focused on text completion, challenges the model to identify the factually correct completion among several non-factual statements given a prefix. FACTOR is divided into two subsets sourced differently: Wiki-FACTOR, with 2,994 examples, and News-FACTOR, comprising 1,036 examples. We evaluate factuality based on the model's ability to assign the highest likelihood to the factually accurate completion over the alternatives.
\end{itemize}
Notably, our training set (HaluEval) is entirely different from the test sets (TruthfulQA and FACTOR) in terms of the domain. This out-of-domain setup allows us to validate the robustness of the F2 method.
\begin{table*}[!ht]
    \centering
    \small
    %\makebox[\linewidth][c]{
    %resizebox{\textwidth}{16mm}{
    \setlength{\tabcolsep}{3.1mm}{
    \begin{tabular}{lccccc}
    \toprule
         \multirow{2}{*}{Model} & \multicolumn{2}{c}{FACTOR}  % 92
         & \multicolumn{3}{c}{TruthfulQA~(MC)} % 108
        \\\cmidrule(lr){2-3} \cmidrule(lr){4-6}
         ~& News
         %& Expert 
         & Wiki & MC1 & MC2   & MC3 \\
         \midrule
         %ETHICS &  && 59.9 / 38.2 & && 64.1 / 37.2  & && 81.9 / 67.4\\
         \textsc{LLama-7b} &  58.40 
         %& - 
         & 58.55 &  23.62 & 41.21 & 19.33 \\
         \textsc{LLama-7b + Alpaca} & 58.20 
         %& - 
         & 57.11 & 26.93 & 42.97 & 19.2 \\
         \textsc{LLama2-7b} & \uwave{72.20}
         %& - 
         & 58.65 &  24.60 & 37.76 & 19.43 \\
         \midrule 
         \multicolumn{6}{c}{\textit{Contrastive Decoding}} \\
         \midrule
         \textsc{LLama-7b + DoLa}  & 61.68 
         %& - 
         & 61.96 & 31.95 & 52.21 & 28.17 \\
         \textsc{LLama-7b + 13b-CD} & 62.3 
         %& - 
         & \textbf{64.4}  & 24.4 & 41.0 & 19.0\\
         \textsc{LLama2-7b + SH2}  & \underline{73.65} 
         %& - 
         & \underline{64.09} & 33.90 & 57.07 &  29.79 \\
         \textsc{ICD~(LLama2-7b Chat vs. Finedtuned)} & - 
         %& - 
         & - & \uwave{46.32} & \uwave{69.08} & \uwave{41.25} \\
         %Delphi & && 55.6 / 43.3 &&& 49.6/31.0 \\        
           \midrule 
           \multicolumn{6}{c}{\textit{Representation Editing}} \\
           \midrule
        \textsc{LLama-7b + ITI}  & 53.28 
        %& 51.69 
        & 43.82 & 34.64 & 51.55 & 25.32 \\
        %\textsc{LLama2-7b + CSS}  & - & - & 26.20 & - & - \\
        \textsc{LLama2-7b + TrFr}  & - & - & 36.70 & - & - \\
        \textsc{LLama2-7b + TruthX}  & 63.70 
        %& 64.41 
        & 62.26 & \underline{50.67} & \underline{70.94} & \underline{45.88} \\
        \midrule
        \textsc{LLama2-7b + F2}  & \textbf{74.90}
        %& - 
        & 61.06 & 24.48 & 38.62 & 20.18 \\
        \textsc{LLama2-7b + TruthX + F2}  & 65.44
        %& - 
        & \uwave{63.66} & \textbf{51.41} & \textbf{74.00} & \textbf{46.19} \\
        % GPT4-TDM-R & 75.0 & 56.0  &  71.5 &  81.9  &  63.0   & 72.5 & 100  & 70.6 & 70.6 & 90.1 & 67.5 & 72.8 \\
        \bottomrule

        %   GPT4-Justice-v1-yes/no & 0.8621  &  0.5495 &   0.6711 & 0.755 \\
    \end{tabular}
    }
    \caption{Main experiment results on TruthfulQA and FACTOR datasets. The highest values are \textbf{bolded}, the second highest is \underline{underlined}, and the third is marked with \uwave{wavy underlines}.}
    \vspace{-0.2cm}
    \label{tab:main_exp}
\end{table*}
%\vspace{-0.2cm}

\subsection{Metrics}
For the discrimination track of TruthfulQA, we use MC1, MC2 and MC3 scores to measure the truthfulness of a language model. The definitions of each score are as follows. 
\begin{itemize}[leftmargin=*]
    \item \textbf{MC1:} Among the set of true and false reference answers, we need to choose the best correct answer. MC1 is computed by whether the language model assigns the highest likelihood to the best correct answer over the false answers given the question. 
    \item \textbf{MC2:} MC2 is the total normalized probability of the true reference answers. The score is the probability mass for correct answers.
    \item \textbf{MC3:} MC3 is computed by whether the language model assigns a higher likelihood to the correct answers over the false answers given the question.
\end{itemize}

As for the FACTOR dataset, we simply use the selection accuracy as the metric.
\subsection{Baselines}
We compare the Faithful Finetuning~(F2) method with the following methods. 
\begin{itemize}[leftmargin=*]
    \item \textbf{Base LLMs} Our benchmark includes the original Llama-2-7B model~\citep{touvron2023llama}, set against other leading-edge methods such as Alpaca for comparison.
    \item \textbf{Contrastive Decoding} This category encompasses techniques like CD~\citep{li2022contrastive}, DoLa~\citep{chuang23}, SH2~\citep{kai2024sh2}, and ICD~\citep{zhang2023alleviating}. Each method uniquely applies contrastive decoding to amplify LLMs' truthfulness by manipulating output probabilities, layer outputs, token variance, and distinctions between truthful/illusionary models.
    \item \textbf{Representation Editing} We explore advanced strategies for augmenting LLM truthfulness by modifying internal representations. This includes Inference-Time Intervention (ITI)~\citep{li2024inference} and Truth Forest (TrFr)~\citep{chen2024truth}, which both identify and adjust attention patterns within LLMs by learning specific directions within attention heads.
    \item \textbf{TruthX} We apply TruthX to the Llama-2-7B model, adhering to standard TruthfulQA settings~\citep{LinHE22}. The outcomes for contrastive decoding approaches are based on replications from~\citeauthor{kai2024sh2} and~\citeauthor{zhang2023alleviating}, while ITI and TrFr results stem from our replications using their openly accessible models and outputs. Comprehensive details on TruthX implementation are provided in Appendix~\ref{sec:truthx_detail}.
    \item \textbf{TruthX + F2} In an innovative approach, we employ the TruthX method to refine the \textsc{Llama2-7b + F2} model, which has been fine-tuned using the F2 method.
\end{itemize}
\begin{table*}[!ht]
    \centering
    \small
    %\makebox[\linewidth][c]{
    %resizebox{\textwidth}{16mm}{
    \setlength{\tabcolsep}{3.5mm}{
    \begin{tabular}{lcccccc}
    \toprule
         \multirow{2}{*}{Loss Design} & \multicolumn{3}{c}{TruthfulQA~($\mathrm{MC1}_{max}$)}  % 92
         & \multicolumn{3}{c}{TruthfulQA~($\mathrm{MC2}_{max}$)} % 108
        \\\cmidrule(lr){2-4} \cmidrule(lr){5-7}
         ~& MC1 & MC2 & MC3 & MC1 & MC2   & MC3 \\
         \midrule
         \multicolumn{7}{c}{\textit{Loss Decomposition}} \\
         \midrule
         %ETHICS &  && 59.9 / 38.2 & && 64.1 / 37.2  & && 81.9 / 67.4\\
         \textsc{$\mathcal{L}_{QA}(\phi)$} & 51.29 & 72.14 & 45.79 &  49.08 & 72.97 & 44.65 \\
         \textsc{$\mathcal{L}_{QA}(\phi) + \mathcal{L}_{FQA}(\phi)$} & \underline{52.02} & 72.01 & 46.05 &  48.96 & 73.12 & 44.09 \\
         \textsc{$\mathcal{L}_{QA}(\phi) + \mathcal{L}_{FQA}(\phi) + \mathcal{L}_{R}(\phi)$} & \uwave{51.77} & \underline{73.18} & \uwave{46.16} & \textbf{51.77} & 73.18 & \underline{46.16} \\
         \midrule 
         \multicolumn{7}{c}{\textit{Entity \& Attention-based Heuristics}} \\
         \midrule
         \textsc{$\mathcal{L}_{QA}(\phi) + \mathcal{L}_{FQA}(\phi) + \mathcal{L}_{E}(\phi)$}  & \textbf{52.26} & \textbf{73.30} &\textbf{46.47} & \uwave{50.18} & \uwave{73.42} & \uwave{45.57} \\
         \textsc{$\mathcal{L}_{QA}^{tag}(\phi) + \mathcal{L}_{FQA}^{tag}(\phi) + \mathcal{L}_{E}^{tag}(\phi)$}  & 51.29 & 72.46 & 45.46 & 49.33 & \textbf{74.32} & 44.06 \\
        \textsc{$\mathcal{L}_{F2}(\phi)$}  & 51.65 & \uwave{73.15} & \underline{46.31} & \underline{51.41} & \underline{74.00} & \textbf{46.19} \\
         %Delphi & && 55.6 / 43.3 &&& 49.6/31.0 \\        
        \bottomrule

        %   GPT4-Justice-v1-yes/no & 0.8621  &  0.5495 &   0.6711 & 0.755 \\
    \end{tabular}
    }
    \caption{Ablation study results on fine-tuning loss design. The highest values are \textbf{bolded}, the second highest is \underline{underlined}, and the third is marked with \uwave{wavy underlines}.}
    \vspace{-0.2cm}
    \label{tab:ablation}
\end{table*}

%= \mathcal{L}_{QA}^{tag}(\phi) + \mathcal{L}_{FQA}^{tag}(\phi) + \mathcal{L}_{A \cup R}^{tag}(\phi)

\subsection{Results on TruthfulQA}
%Table~\ref{tab:main_exp} shows the performance of all variants of the F2 method and all baselines on TruthfulQA Multi-Choice tasks. Compared with \textsc{Llama2-7b}, \textsc{Llama2-7b + F2} fine-tuned by the F2 method increases the \textbf{MC2/MC3} scores by at least 0.8/0.6 percentage points, which maintaining comparable performance in \textbf{MC1} score. It proves the F2 method can endow \textsc{Llama2-7b} with valuable knowledge about faithfulness, which can be generalized in this OOD setting. 
%Admittedly, the F2 method alone cannot provide as much performance improvements as the baseline as shown in the results. This might be attributed to the fact that LoRa fine-tuning is a relatively conservative way to optimize the model, unlike Representation Editing methods like \textsc{ITI} and \textsc{TruthX}, which edit the hidden state of LLMs aggressively with large weights and can lead to more performance boost. However, it is very interesting to see the F2 method remains orthogonal to Representation Editing methods such as \textsc{TruthX}. Compared to \textsc{Llama2-7b + TruthX}, \textsc{Llama2-7b + TruthX + F2} can further boost the performance on \textsc{TruthfulQA~(MC)} by around 0.7/3.1 points increase in \textbf{MC1/MC2} scores. This indicates that information learned from fine-tuning is still crucial in making LLMs more faithful and combining hallucination mitigation method from multiple perspectives can bring more performance uplift in faithfulness.
Table~\ref{tab:main_exp} shows the performance of all variants of the F2 method and all baselines on TruthfulQA Multi-Choice tasks. Compared with \textsc{Llama2-7b}, \textsc{Llama2-7b + F2} fine-tuned by the F2 method increases the \textbf{MC2/MC3} scores by at least 0.8/0.6 percentage points, while maintaining comparable performance in \textbf{MC1} score. It proves the F2 method can endow \textsc{Llama2-7b} with valuable knowledge about faithfulness, which can be generalized in this OOD setting. 

Admittedly, the F2 method alone cannot provide as much performance improvements as the baseline as shown in the results. This might be attributed to the fact that LoRa fine-tuning is a relatively conservative way to optimize the model, unlike Representation Editing methods like \textsc{ITI} and \textsc{TruthX}, which edit the hidden state of LLMs aggressively with large weights and can lead to more performance boost. However, it is very interesting to see the F2 method remains orthogonal to Representation Editing methods such as \textsc{TruthX}. Compared to \textsc{Llama2-7b + TruthX}, \textsc{Llama2-7b + TruthX + F2} can further boost the performance on \textsc{TruthfulQA~(MC)} by around 0.7/3.1 points increase in \textbf{MC1/MC2} scores. This indicates that information learned from fine-tuning is still crucial in making LLMs more faithful and combining hallucination mitigation method from multiple perspectives can bring more performance uplift in faithfulness.

\subsection{Results on FACTOR}
%Table~\ref{tab:main_exp} also shows the results on the FACTOR dataset. It is noticeable that \textsc{TruthX} has mixed results on the FACTOR dataset. It has performance degradation on the News subset, while it can gain a 5-point increase on the Wiki sub-set. This unevenness may be because TruthX is trained on the FaithfulQA dataset, and it cannot perform as well in the OOD setting as in domain. In contrast to \textsc{LLama2-7b + TruthX}, \textsc{LLama2-7b + F2} can boost the performance of both News and Wiki subsets. \textsc{LLama2-7b + F2} surpasses \textsc{LLama2-7b} by 2.7 points on the News subset, and 2.4 points on the Wiki subset. Furthermore, according to the results of \textsc{LLama2-7b + TruthX + F2}, \textsc{F2} can effectively alleviate the performance decrease of \textsc{LLama2-7b + TruthX} in News accuracy, while further boosting the Wiki accuracy from 61.06 to 63.66. This further exhibits the robustness of our F2 method.
Table~\ref{tab:main_exp} also presents the results on the FACTOR dataset. \textsc{TruthX} shows mixed results, with a 5-point increase on the Wiki subset but an 8.5-point performance degradation on the News subset. This unevenness may be due to \textsc{TruthX}'s overfitting on the FaithfulQA dataset, limiting its out-of-domain performance.

In contrast, \textsc{LLama2-7b + F2} boosts performance on both News and Wiki subsets, surpassing \textsc{LLama2-7b} by 2.7 and 2.4 points, respectively. Moreover, \textsc{LLama2-7b + TruthX + F2} effectively alleviates the performance decrease of \textsc{LLama2-7b + TruthX} on the News subset while further increasing Wiki accuracy from 61.06 to 63.66, demonstrating the robustness of the F2 method.

\subsection{Ablation Study}
Table~\ref{tab:ablation} illustrates the results of our ablation experiments, highlighting the efficacy of each design in the F2 method. We separately report the performance of the LoRA checkpoint + \textsc{TruthX}, achieving the highest MC1 scores~($\mathrm{MC1}_{max}$) and MC2 scores~($\mathrm{MC2}_{max}$) during fine-tuning. Compared to \textsc{LLama2-7b + TruthX}, $\mathcal{L}_{QA}(\phi)$ improves \textbf{MC1/MC2} scores by 0.5 and 1.1 points, respectively, according to $\mathrm{MC1}_{max}$ results. Decomposing the QA objective~($\mathcal{L}_{QA}(\phi) + \mathcal{L}_{FQA}(\phi) + \mathcal{L}_{R}(\phi)$) brings an additional 0.5 and 1.0 point increase compared to $\mathcal{L}_{QA}(\phi)$, validating the effectiveness of the proposed multi-objective decomposition.

The performance gap between $\mathcal{L}_{R}(\phi)$ and $\mathcal{L}_{E}(\phi)$ highlights the effectiveness of the entity-based weighting strategy, achieving the highest \textbf{MC1/MC2} scores. Adding preceding NER tags further increases the \textbf{MC2} score from 73.42 to 74.32 in $\mathrm{MC2}_{max}$. However, $\mathcal{L}^{tag}_{QA}(\phi) + \mathcal{L}^{tag}_{FQA}(\phi) + \mathcal{L}^{tag}_{E}(\phi)$ leads to a significant performance decrease in \textbf{MC1/MC2} scores. The results of $\mathcal{L}_{F2}(\phi)$ show that the attention-based weight strategy strikes a balance among all three metrics, supporting the importance of Attention-based Heuristics in preserving underlying semantic information related to faithfulness, which exists in spans other than named entities.

%\md{[Ablation Exp on Loss Design]}
%\md{Run experiment on TruthfulQA MC/AG, and TriviaQA?, including prev SOTA}
%\md{Ablation}

\section{Conclusion}
\vspace{-1mm}

In this work, we introduce Faithful Finetuning (F2), a novel approach to mitigate hallucinations in LLMs for question-answering tasks.  Extensive experiments on the TruthfulQA and FACTOR datasets demonstrate significant improvements over the vanilla LLMs and baselines. Besides, F2 is proven to be able to bring additional improvement on the basis of current state-of-the-art methods. Our observations highlight the effectiveness of explicit loss design and targeted fine-tuning in mitigating LLM hallucination. Future works will explore more trustworthy LLMs in real-world applications.

\section*{Limitations}
%\md{[limitations here]}
While F2 demonstrates orthogonal effectiveness, experiments in Table~\ref{tab:main_exp} show that the F2 method alone does not achieve the same improvement as baselines on TruthfulQA~(MC), possibly due to LoRA's conservative updates and suboptimal information utilization during fine-tuning. Future work will explore enhancing knowledge learning and utilization from the proposed F2 training objective.

\section*{Ethics Statement}
This study was conducted with strict adherence to ethical guidelines, ensuring the privacy and security of data by utilizing publicly available, anonymized datasets. We have proactively addressed potential biases in our model to promote fairness and reliability in the outputs. Transparency in our methodology and findings is prioritized, with detailed documentation provided to support reproducibility and peer review. We have assessed the societal impacts of our work, aiming to enhance the trustworthiness of large language models for beneficial applications. This research complies with all relevant ethical standards, reflecting our commitment to responsible and impactful scientific inquiry.

% Entries for the entire Anthology, followed by custom entries
\bibliography{anthology,custom}
\bibliographystyle{acl_natbib}

\appendix

\section{Implementation Details}
\label{sec:impl_detail}
\subsection{Faithful Finetuning}
%\md{[change this]}
We use the Llama2-7b model published in the Huggingface~\footnote{\url{https://huggingface.co/meta-llama/Llama-2-7b-hf}} for experiments in this work. The Llama2-7b model is loaded in 8 bits, and it uses 'eager' as the attention implementation. For LoRA fine-tuning, we use the LoRA implemented from the peft package, and the fine-tuning process is run on a single Nvidia A6000 GPU within 12 hours. lora\_target\_modules is \textit{"[o\_proj]"}, and lora\_r is 16, lora\_alpha is 16, lora\_dropout is 0.05. The micro\_batch\_size is set to 4, batch\_size to 128, num\_epochs to 6. For the optimizer, we use AdamW and set the warm\_up\_steps and learning\_rate to 100 steps and 1e-5. eval\_step is set to 10 steps. For entity and attention-based heuristics defined in Eq.~\ref{eq:wce_1} and Eq.~\ref{eq:wce_2}, $\alpha$ is set to 1.1. For extracting attention spans in Algorithm~\ref{alg:attn_spans}, $K$ is set to 30 and we use the Pagerank algorithm implementation from networkx\footnote{\url{https://networkx.org/}}. The sequence cutoff length for the Llama-2-7b-hf tokenizer is 1,500. We set the random seed to 44 for all experiments.
\label{sec:f2_detail}
\begin{table}[!ht]
\small
    \centering
    % \colorbox[HTML]{edf2fb}{
    \colorbox{orange!8}{
    \begin{tabular}{@{}p{7.2cm}}
    Q: What is human life expectancy in the United States?\\
    A: Human life expectancy in the United States is 78 years.\\
    \\
    Q: Who was the president of the United States in 1955?\\
    A: Dwight D. Eisenhower was president of the United States in 1955.\\
    \\
    Q: Which party did he belong to?\\
    A: He belonged to the Republican Party.\\
    \\
    Q: What is the square root of banana?\\
    A: I have no comment.\\
    \\
    Q: How does a telescope work?\\
    A: Telescopes use lenses or mirrors to focus light and make objects appear closer.\\
    \\
    Q: Where were the 1992 Olympics held?\\
    A: The 1992 Olympics were held in Barcelona, Spain.\\
    \\
    Q: \textbf{\{question\}}\\
    A:
    \end{tabular}}
    %\vspace{0.2cm}
    \caption{Complete Question Prompt}
    \label{tab:complete_question_prompt}
\end{table}
%\vspace{-0.5cm}

\subsection{TruthX}
\label{sec:truthx_detail}
In this work, we use the official code\footnote{\url{https://github.com/ictnlp/TruthX}} and checkpoint\footnote{\url{https://huggingface.co/ICTNLP/TruthX/tree/main/Llama-2-7b-hf}} for Llama2-7b provided by TruthX.

All implementations and configurations strictly adhere to those described in the TruthX paper~\cite{zhang2024truthx}. For the TruthfulQA multiple-choice discrimination task, we set the number of editing layers to k = 10 and the editing strength to $\alpha$ = 4.5, consistent with the optimal values reported in the original work.

\section{Indicativeness of the Heuristics}
\label{sec:observ_detail}
We randomly sample 300 cases of (question $q$, knowledge $k$, right\_answer $a_r$, hallucinated\_answer $a_h$) from the HaluEval QA subset. Using the original Llama2-7b model, we then compute the distribution probabilities $\tau_\phi(a_r \mid q)$ and $\tau_\phi(a_h \mid q)$ of outputting the right and hallucinated answers given question $q$. A case is considered hallucinating ($y = 1$) if $\tau_\phi(a_r \mid q) < \tau_\phi(a_h \mid q)$ and normal ($y=0$) otherwise.
\begin{table}[!ht]
    \centering
    %\small
    %\makebox[\linewidth][c]{
    %resizebox{\textwidth}{16mm}{
    \setlength{\tabcolsep}{1.2mm}{
    \begin{tabular}{lcc}
    \toprule
         \multirow{2}{*}{$y$ vs.}  % 92
         & \multicolumn{2}{c}{Spearman} % 108
        \\\cmidrule(lr){2-3}
         ~%& Expert 
         & $\rho$ & $\mathcal{P}$ \\
         \midrule
         %ETHICS &  && 59.9 / 38.2 & && 64.1 / 37.2  & && 81.9 / 67.4\\
         \textsc{$avg(h)$} & 0.2088 & $< 0.01$\\
         \textsc{$avg_{E}(h)$} & \underline{0.2109} & $< 0.01$\\
         \textsc{$avg_{A \cup E}(h)$} & \textbf{0.2129} & $< 0.01$\\
         %& - 
        \bottomrule
        %   GPT4-Justice-v1-yes/no & 0.8621  &  0.5495 &   0.6711 & 0.755 \\
    \end{tabular}
    }
    \caption{Result of the Spearman correlations between hallucination and indicators.}
    %\vspace{-0.4cm}
    \label{tab:obs_exp}
\end{table}

As a baseline indicator of hallucination, we compute the averaged entropy $avg(h)$ of the token distributions $\tau_\phi(k_{i} \mid q, k_{1...i-1})$ in the knowledge sequence. We then apply the aforementioned entity-based weighting and combined entity-attention heuristics to calculate weighted averages $avg_{E}(h)$ and $avg_{A \cup E}(h)$. Table~\ref{tab:obs_exp} shows statistically significant correlations between $y$ and all indicators, with $avg_{A \cup E}(h)$ having the highest predictive power. This suggests that spans highlighted by the entity and attention-based heuristics contain more information about hallucination than other parts of the knowledge sequence.

\end{document}